\documentclass[10pt,twocolumn,letterpaper]{article}

\usepackage[pagenumbers]{cvpr}

\usepackage{multirow}

\definecolor{cvprblue}{rgb}{0.21,0.49,0.74}
\usepackage[breaklinks,colorlinks,allcolors=cvprblue]{hyperref}
\hypersetup{
  pdftitle={CHAMP: Coordinated Humanoid Loco-Manipulation with Whole-Body Motion Priors},
  pdfauthor={Lijun Han, Muyuan Ma, Zichen Wang, Siyuan Liu, Yuxi Liu, Yutong Wang, Bohao Zhang, Heng Zhou, Wei Sui}
}

\title{CHAMP: Coordinated Humanoid Loco-Manipulation\\ with Whole-Body Motion Priors}

\author{
Lijun Han\textsuperscript{1,*}\quad
Muyuan Ma\textsuperscript{2,*}\quad
Zichen Wang\textsuperscript{1,*}\quad
Siyuan Liu\textsuperscript{2,*}\\[3pt]
Yuxi Liu\textsuperscript{1}\quad
Yutong Wang\textsuperscript{1}\quad
Bohao Zhang\textsuperscript{1}\quad
Heng Zhou\textsuperscript{1}\quad
Wei Sui\textsuperscript{1,\textdagger}\\[7pt]
{\normalsize \textsuperscript{1}D-Robotics}\\[2pt]
{\normalsize \textsuperscript{2}Institute of Automation, Chinese Academy of Sciences}\\[3pt]
{\small \textsuperscript{*}Equal contribution.\quad
\textsuperscript{\textdagger}Corresponding author.}\\[4pt]
{\small \textbf{Project page:} \url{https://han-li-jun.github.io/CHAMP/index.html}}
}
\date{}

\makeatletter
\apptocmd{\@maketitle}{\noindent\begin{minipage}{\textwidth}
  \centering
  \includegraphics[width=\linewidth]{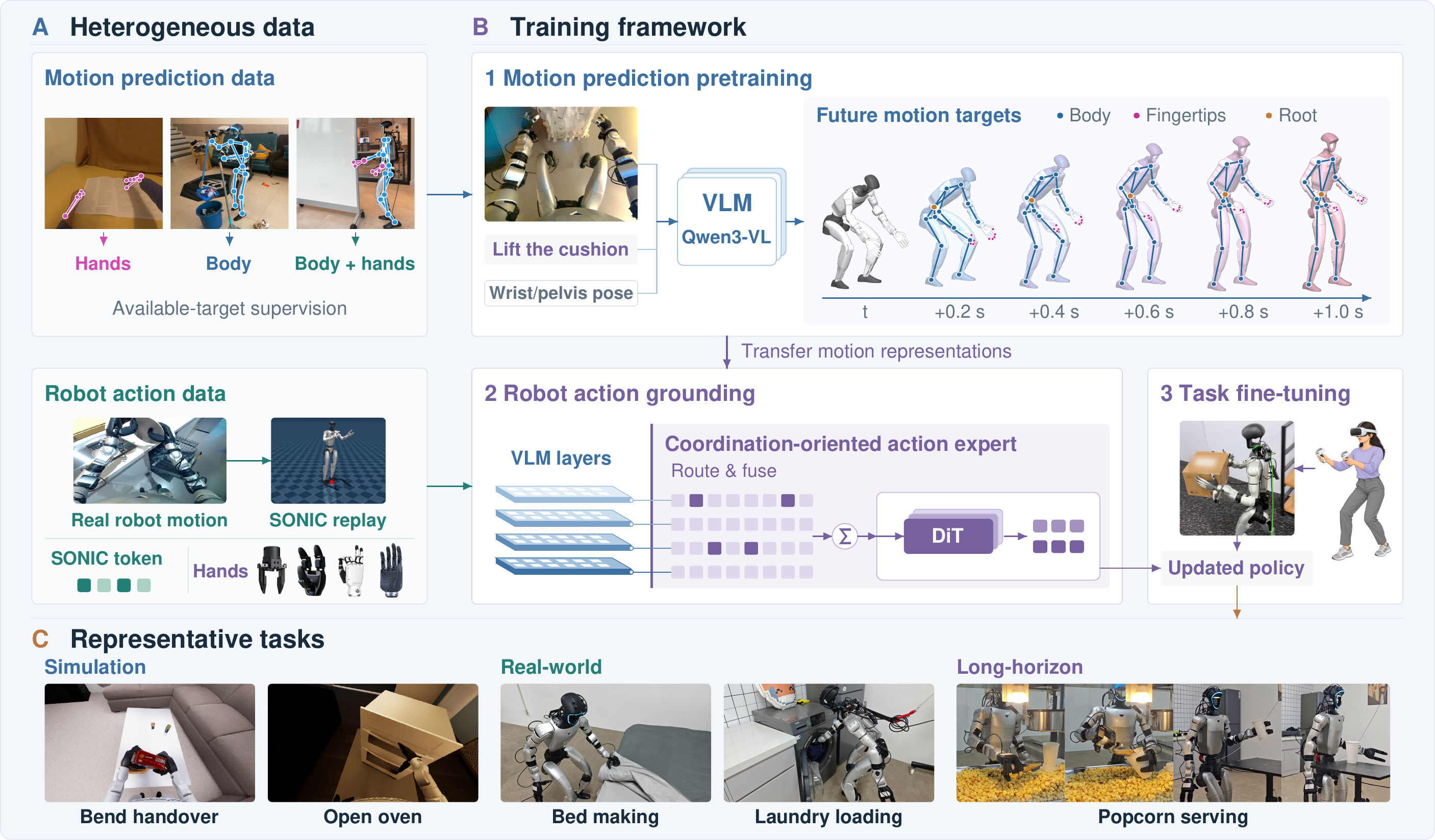}
  \captionsetup{hypcap=false,skip=6pt}
  \captionof{figure}{\textbf{Overview of CHAMP.}
  CHAMP pretrains a VLM on heterogeneous human and robot data to predict future body and hand motion.
  The Coordination-oriented Action Expert adaptively extracts and fuses multi-layer features for robot action grounding, followed by task-specific fine-tuning.
  Representative tasks span simulation and real-world settings, including long-horizon popcorn serving.}
  \label{fig:framework}
\end{minipage}\par
\vspace{12pt}
}{}{%
  \PackageError{paper}{Could not insert the framework figure}{Check the title-block hook.}%
}
\makeatother

\begin{document}
\maketitle
\begin{abstract}
Coordinating locomotion and dexterous manipulation is fundamental to humanoid robots operating in human environments.
A key challenge is learning motion priors from diverse data and using them effectively for coordinated robot control.
We present CHAMP, a unified framework for humanoid loco-manipulation that combines whole-body motion pretraining with adaptive feature extraction.
CHAMP pretrains a vision-language model to predict future body and hand motion from approximately 2,298 hours of heterogeneous human and robot data.
A shared motion representation enables unified whole-body motion learning across embodiments from datasets with body annotations, hand annotations, or both.
To ground these motion priors in robot control, we introduce a Coordination-oriented Action Expert (CAE) and train it on replay data generated using the target controller.
Through sparse routing, CAE adaptively selects, transforms, and fuses features from multiple layers of the pretrained model to jointly generate body and hand actions.
In experiments, CHAMP achieves overall success rates of 84.8\% on seven simulated tasks and 91.4\% on seven real-world tasks with a Unitree G1, outperforming the evaluated baselines in both settings.
In particular, CHAMP achieves 70.0\% success on the long-horizon popcorn-serving task, compared with 40.0\% for the best-performing baseline.
\end{abstract}

\section{Introduction}
\label{sec:intro}

Humanoid robots hold the promise of assisting people across a wide range of everyday activities~\cite{hu2026workhorse,jiang2025wholebodyvla,shi2026egohumanoid}.
Fulfilling this promise requires a broad repertoire of motor skills.
A key capability is loco-manipulation, which integrates locomotion and object manipulation through whole-body coordination~\cite{he2025viral,li2026wholebodywam_wbc}.
Locomotion and manipulation are tightly coupled.
Body motion enables the hands to reach and interact with objects, while the demands of manipulation shape how the robot moves and maintains balance~\cite{li2026coordex}.
These dependencies evolve throughout execution, requiring a policy to coordinate body and hand motion over time.
We therefore aim to enable coordinated humanoid loco-manipulation by learning whole-body motion priors that capture these relationships and integrating them into robot policies.

Learning such priors requires diverse examples of body and hand motion, yet collecting demonstrations on physical humanoids is costly and difficult to scale~\cite{zheng2026egoscale,pan2026ironmind}.
Existing human and robot motion datasets offer a broader source of supervision beyond demonstrations collected for a target robot~\cite{lin2023motionx,xperience_10m,fan2023arctic}.
$\Psi_0$ pretrains a vision-language model (VLM) to predict wrist poses and fingertip positions~\cite{wei2026psi0}, while WholeBodyWAM learns body-motion priors by predicting the future rotations of 21 body joints~\cite{zhang2026wholebodywam_motion}.
Despite these advances, these methods do not jointly model body motion and fine-grained hand motion during pretraining.
This motivates us to learn unified whole-body motion priors that explicitly capture body--hand coordination through joint motion prediction, using heterogeneous datasets with annotations for body motion, hand motion, or both.

Translating these whole-body motion priors into coordinated robot behavior requires adapting representations learned for motion prediction to robot action generation.
$\Psi_0$ conditions its action expert on final-layer VLM features~\cite{wei2026psi0}.
GR00T N1.6~\cite{nvidia2025grootn16} and N1.7~\cite{nvidia2026grootn17} instead use intermediate VLM features.
DiT4DiT~\cite{ma2026dit4dit} and MotionWAM~\cite{zheng2026motionwam} use features from a fixed video-model layer for action generation, with DiT4DiT reporting better performance using intermediate rather than final-layer features.
These methods use features from a source layer selected in advance.
However, as the task or observation changes, the features most suitable for robot control may come from different layers.
We therefore investigate whether jointly learning which layers to read and how to adapt their features, under robot-action supervision, can better support coordinated whole-body control.

To achieve coordinated whole-body control for humanoid loco-manipulation, we present \textbf{CHAMP}, a unified framework that combines whole-body motion pretraining with adaptive motion feature extraction (Fig.~\ref{fig:framework}).
First, we pretrain a VLM to predict future body and hand motion from heterogeneous human and robot data, supervising only the motion targets available in each sample.
Second, we use SONIC-replayed robot trajectories~\cite{luo2025sonic} to ground the learned representations in robot control.
In this stage, we introduce a \textbf{Coordination-oriented Action Expert (CAE)} that adaptively selects and fuses features from the pretrained VLM.
Through joint layer--expert routing, CAE uses the fused features to condition flow-matching generation of SONIC motion latents and hand actions.
Finally, we fine-tune the policy with task-specific demonstrations collected through whole-body teleoperation.
Our contributions are threefold:
\begin{itemize}
  \item We propose CHAMP, a unified framework for coordinated humanoid loco-manipulation.
  It learns whole-body motion priors from heterogeneous human and robot motion data and transfers the learned priors to robot control.
  \item We introduce CAE to adaptively extract motion features from the VLM pretrained for whole-body motion prediction.
  It jointly selects source layers and projection experts for each input token, transforming and fusing these features for coordinated whole-body control.
  \item Experiments demonstrate CHAMP's effectiveness for coordinated humanoid loco-manipulation, including long-horizon execution.
  CHAMP achieves overall success rates of 84.8\% on seven simulated tasks and 91.4\% on seven real-world tasks, outperforming the evaluated baselines in both settings.
\end{itemize}

\begin{figure*}[t]
  \centering
  \includegraphics[width=\textwidth]{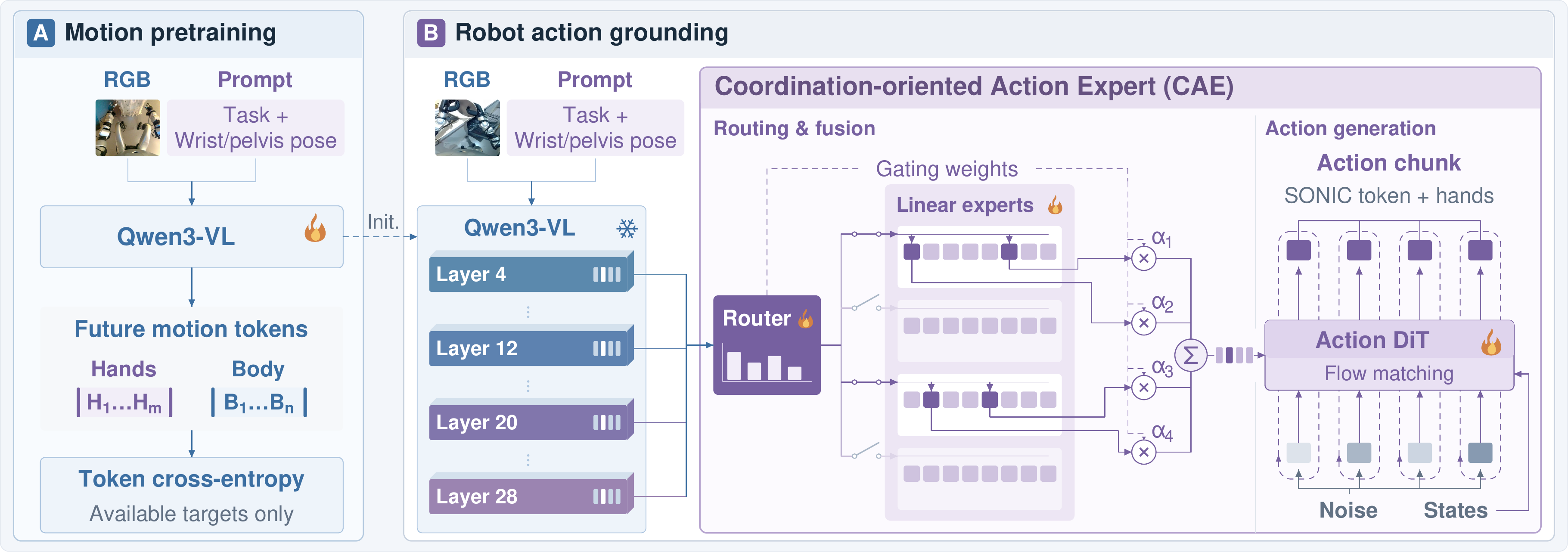}
  \caption{\textbf{Model architecture.}
  (A) Qwen3-VL is pretrained to predict future body and hand motion tokens using available annotations.
  (B) Robot action grounding freezes the VLM and trains CAE, which adaptively selects, projects, and fuses multi-layer VLM features through sparse layer--expert routing.
  The fused features and robot state condition a flow-matching DiT for coordinated whole-body control.}
  \label{fig:model_architecture}
\end{figure*}

\section{Related Work}
\label{sec:related}

\subsection{Humanoid Loco-Manipulation}
Humanoid loco-manipulation requires coordinated locomotion, posture adjustment, and object interaction.
WholeBodyVLA~\cite{jiang2025wholebodyvla} and $\Psi_0$~\cite{wei2026psi0} use high-level VLA policies to generate upper-body actions together with locomotion and posture commands, while low-level RL controllers handle lower-body execution.
This division leaves detailed leg motion to the locomotion controller, limiting how directly the task policy can specify lower-body behavior.
CoorDex~\cite{li2026coordex} learns separate body and hand control priors and coordinates them through reinforcement learning with task-specific rewards.
Motion-tracking models such as SONIC~\cite{luo2025sonic} provide a reusable foundation for whole-body execution, supporting diverse movements through a shared controller.
Building on this foundation, OpenHLM~\cite{hu2026openhlm} adapts a VLA to generate full-body joint targets tracked by SONIC.
WholeBodyWAM~\cite{li2026wholebodywam_wbc} connects pretrained world--action models to whole-body controllers through a unified interface.
To achieve more coordinated whole-body control for humanoid loco-manipulation, CHAMP focuses on learning unified body and hand motion priors and adaptively extracting features from the motion-pretrained VLM for robot control.

\subsection{Learning Whole-Body Motion Priors}
Human and robot motion datasets provide heterogeneous resources for learning motion priors, with different embodiments, motion coverage, and annotations.
Human datasets such as Motion-X~\cite{lin2023motionx}, EgoDex~\cite{hoque2026egodex}, ARCTIC~\cite{fan2023arctic}, and Xperience-10M~\cite{xperience_10m} provide annotations of body motion, hand motion, or both.
Robot datasets such as HIW-500~\cite{hiw500_2026} and Humanoid Everyday~\cite{zhao2025humanoideveryday} contain real-world humanoid demonstrations of locomotion and manipulation.
$\Psi_0$~\cite{wei2026psi0} and $\lambda_0$~\cite{lambda2026lambda0} learn motion priors by predicting future hand motion during pretraining and subsequently ground these priors in robot control.
WholeBodyWAM~\cite{zhang2026wholebodywam_motion} learns body-motion priors by predicting future rotations of 21 body joints.
WB-WAM~\cite{qin2026wbwam} learns whole-body motion priors from heterogeneous body and hand data in a robot-specific action space.
In contrast, CHAMP jointly predicts 3D body-joint and fingertip positions, together with root motion, in a shared motion representation across humans and multiple robot embodiments.
These shared targets support motion-prior learning without requiring all sources to be mapped to a single robot's joint-angle space.

\subsection{Pretrained Representations for Robot Control}
The interface between a pretrained backbone and an action policy determines how learned knowledge supports control.
GR00T models such as N1.6~\cite{nvidia2025grootn16} and N1.7~\cite{nvidia2026grootn17} extract features from a fixed intermediate VLM layer.
DiT4DiT~\cite{ma2026dit4dit} and MotionWAM~\cite{zheng2026motionwam} condition action generation on features extracted from fixed video-model depths.
Although these methods use attention to combine features, they extract those features from a fixed layer.
$\omega$-0~\cite{li2026omega0} extracts a summary feature from the final VLM layer.
Motion and video queries then integrate this summary with additional visual and language features.
WholeBodyWAM~\cite{zhang2026wholebodywam_motion} models video, motion, and actions through three branches that interact via attention at each transformer layer.
CAE extracts features adaptively from the motion-pretrained VLM.
For each input token, it jointly selects source layers and layer-specific projection experts.
By combining this adaptive selection with feature projection and fusion, the policy learns both where to access motion knowledge and how to adapt it for whole-body control.

\section{Method}
\label{sec:method}

To enable coordinated humanoid loco-manipulation, CHAMP combines whole-body motion pretraining with adaptive extraction of motion features.
As illustrated in Figure~\ref{fig:model_architecture}, a VLM learns to predict body and hand motion from heterogeneous data, and CAE adaptively extracts its internal features for coordinated body and hand action generation.
We first define the control task (Sec.~\ref{sec:overview}), then describe motion prediction pretraining (Sec.~\ref{sec:motion_pretraining}), robot action grounding (Sec.~\ref{sec:action_grounding}), and task fine-tuning (Sec.~\ref{sec:task_finetuning}).

\subsection{Problem Formulation}
\label{sec:overview}
We formulate humanoid loco-manipulation as the generation of coordinated body and hand actions conditioned on multimodal inputs.
At time $t$, the policy receives a single RGB image $I_t$, a language instruction $\ell$, the poses of both wrists and the pelvis (collectively denoted by $\mathbf{p}_t$), and body and hand joint angles $\mathbf{s}_t$.
The VLM receives $I_t$ and a text prompt encoding $\ell$ and $\mathbf{p}_t$, while $\mathbf{s}_t$ separately conditions the action expert.

Each action ${a}_t$ includes a SONIC motion latent $\mathbf{z}_t$ and commands $\mathbf{u}_t^{\mathrm{L}}$ and $\mathbf{u}_t^{\mathrm{R}}$ for the left and right end effectors:
\begin{equation}
  \begin{aligned}
    \mathbf{a}_t &= [\mathbf{z}_t;\mathbf{u}_t^{\mathrm{L}};\mathbf{u}_t^{\mathrm{R}}]\in\mathbb{R}^{d_a},\\
    \mathbf{A}_t &= [\mathbf{a}_t,\ldots,\mathbf{a}_{t+T_a-1}]^{\mathsf{T}}
      \in\mathbb{R}^{T_a\times d_a},
  \end{aligned}
  \label{eq:joint_action_chunk}
\end{equation}
where $T_a$ is the action horizon and $d_a$ is the action dimension.
The policy, parameterized by $\theta$, predicts this action chunk as
\begin{equation}
  \mathbf{A}_t \sim \pi_\theta(\,\cdot\mid I_t,\ell,\mathbf{p}_t,\mathbf{s}_t).
  \label{eq:policy_formulation}
\end{equation}

\subsection{Heterogeneous Motion Prediction Pretraining}
\label{sec:motion_pretraining}
We pretrain Qwen3-VL~\cite{bai2025qwen3vl} to predict future body and hand motion (Fig.~\ref{fig:model_architecture}A), using heterogeneous human and robot data.
Shared motion targets let datasets with different embodiments and annotations contribute to the same VLM, without requiring a common robot joint-angle space.

\paragraph{Pretraining data.}
We use heterogeneous motion data from humans, real robots, and simulation for pretraining.
Human data come from EgoDex~\cite{hoque2026egodex}, Xperience-10M~\cite{xperience_10m}, and ARCTIC~\cite{fan2023arctic}.
EgoDex provides hand-only annotations, while Xperience-10M and ARCTIC provide both body and hand annotations.
Real-robot data come from HIW-500~\cite{hiw500_2026}, PSI-Real~\cite{wei2026psi0}, RoboCOIN~\cite{wu2025robocoin}, HumanoidEveryday~\cite{zhao2025humanoideveryday}, and UnifoLM-WBT~\cite{unitree2026unifolmwbt}.
HIW-500 provides body-only annotations because its robots use grippers.
PSI-Real and the RoboCOIN subset recorded on Unitree G1 robots with Dex3 hands provide hand-only annotations because their recordings lack complete leg-joint states.
HumanoidEveryday provides body and hand annotations from Unitree G1 and H1 robots with dexterous hands.
UnifoLM-WBT includes Unitree G1 recordings with different end effectors.
Its Dex1 gripper subset provides body-only annotations, while the BrainCo and Inspire dexterous-hand subsets provide both body and hand annotations.
GR00T-X-Embodiment-Sim~\cite{nvidia2025grootxsim} provides body and hand annotations from simulated Fourier GR1 robots equipped with dexterous hands.

\begin{figure}[t]
  \centering
  \includegraphics[width=\columnwidth]{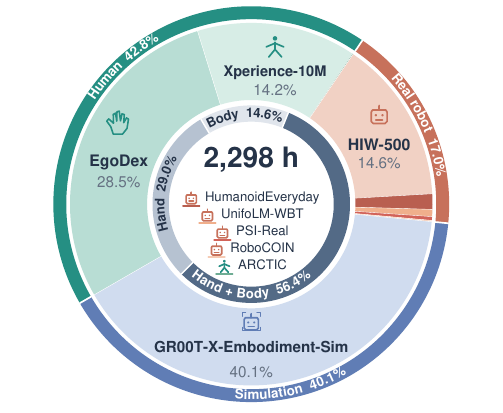}
  \caption{\textbf{Motion pretraining data.}
  The outer and middle rings show data domains and sources.
  The inner ring independently summarizes available motion annotations, without assigning its sectors to individual sources.}
  \label{fig:pretraining_data}
\end{figure}

Together, these nine datasets provide approximately 2,298 hours of motion data.
Figure~\ref{fig:pretraining_data} shows that human, real-robot, and simulation data account for 42.8\%, 17.0\%, and 40.1\% of the total duration, respectively.
Both body and hand annotations are available for 56.4\% of the data, while 29.0\% have hand-only annotations and 14.6\% have body-only annotations.

\paragraph{Motion representation and prediction.}
Humans and different humanoid robot embodiments have key body parts in common, including the pelvis, shoulders, elbows, wrists, hips, knees, and ankles.
For embodiments with dexterous hands, wrist and fingertip motion can likewise describe hand movement.
We therefore use the motion of these shared body and hand structures to define a common motion representation across embodiments.
Body motion $\mathbf{m}^{\mathrm{body}}$ comprises the pelvis position and orientation, together with the 3D positions of 12 body joints, six on each side.
Hand motion $\mathbf{m}^{\mathrm{hand}}$ comprises the positions and orientations of both wrists, together with fingertip positions per hand.

Each pretraining sample contains a single RGB image $I_t$, an instruction $\ell$, and current pose information $\mathbf{p}_t$.
Let $\mathcal{C}_t$ denote the sample's annotated motion types, which may be body motion, hand motion, or both.
The corresponding pose input $\mathbf{p}_t$ contains the pelvis pose, wrist poses, or both, respectively.
For each $c\in\mathcal{C}_t$, we define the future motion sequence
\begin{equation}
  \mathbf{M}_t^c = (\mathbf{m}_{t+1}^c,\ldots,\mathbf{m}_{t+T_m}^c),
  \label{eq:available_motion_targets}
\end{equation}
where each prediction step corresponds to 0.2 seconds and $T_m=5$ steps cover the next 1 second.

We express body-joint positions relative to the pelvis and fingertip positions relative to the corresponding wrist.
The pelvis and wrist positions and orientations are expressed in the same reference frame within each trajectory.
We express future positions and orientations relative to their current values, encode rotations using a 6D representation, and normalize the resulting motion sequences before tokenization.
Source-specific coordinate transformations and normalization details are provided in the supplementary material (Secs.~\ref{supp:motion_data} and~\ref{supp:motion_tokenization}).

We train separate FAST tokenizers~\cite{pertsch2025fast} for body and hand motion, each with its own vocabulary of 8,192 tokens.
These tokenizers encode the motion sequences $\mathbf{M}_t^c$ into a token sequence $\mathbf{y}_t=(y_{t,1},\ldots,y_{t,N_t})$, where $N_t$ is the number of tokens.

The VLM is trained to autoregressively predict this token sequence $\mathbf{y}_t$ conditioned on $I_t$, $\ell$, and $\mathbf{p}_t$.
For a minibatch $\mathcal{B}$, we optimize the token-level cross-entropy loss
\begin{equation}
 \mathcal{L}_{\mathrm{motion}}=
 -\frac{\displaystyle\sum_{t\in\mathcal{B}}\sum_{n=1}^{N_t}
 \log p_\phi(y_{t,n}\mid\mathbf{y}_{t,<n},I_t,\ell,\mathbf{p}_t)}
 {\displaystyle\sum_{t\in\mathcal{B}}N_t},
 \label{eq:motion_pretraining_loss}
\end{equation}
where $\phi$ denotes the VLM parameters and $\mathbf{y}_{t,<n}$ denotes the first $n-1$ tokens of $\mathbf{y}_t$.
We fine-tune all 2.16B VLM parameters using this objective for 250,000 steps with a global batch size of 768.
The learning rate is $10^{-4}$ for the language model and $10^{-5}$ for the visual modules.
Training takes approximately 150 hours on 48 NVIDIA H20 GPUs.

\subsection{Robot Action Grounding}
\label{sec:action_grounding}

\paragraph{Robot action data.}
To ground the pretrained motion prior in the control of a target robot, we select robot datasets whose body trajectories can be tracked by that robot.
We use Unitree G1 as the target platform.
The data come from HIW-500~\cite{hiw500_2026}, HumanoidEveryday~\cite{zhao2025humanoideveryday}, and UnifoLM datasets, spanning grippers and dexterous hands.
We use SONIC's motion encoder~\cite{luo2025sonic} to convert the recorded body trajectories into motion latents.
Using these latents, we replay the body motions on a simulated G1 with the SONIC controller to screen for instability and excessive drift.
After quality filtering, we retain 40,016 trajectory segments totaling approximately 254.5 hours.
Each action in a retained segment contains a 64-dimensional motion latent and the corresponding end-effector commands.
The resulting action sequences are paired with RGB observations, instructions, and robot states.
Motion encoding, simulation replay, and data filtering are detailed in Sec.~\ref{supp:sonic_replay}.

\paragraph{Coordination-oriented Action Expert.} To generate coordinated robot actions from the pretrained motion representations, we freeze the VLM and train CAE on replayed robot data (Fig.~\ref{fig:model_architecture}B).
CAE combines a sparse mixture-of-experts (MoE) module for cross-layer feature projection and fusion with a diffusion transformer (DiT) for robot action generation. Unlike conventional MoE, whose experts process features from the same depth, our MoE associates each expert group with a distinct VLM depth. For each token, a joint router selects layer--expert pairs, determining both which motion representations to use and how to transform and fuse them into the DiT hidden space.
CAE has approximately 589M trainable parameters.

The MoE reads features from $N$ candidate layers $\mathcal{L}=\{r_1,\ldots,r_N\}$, each paired with $E$ affine projection experts that map the layer's features into the DiT hidden space.
For each VLM input token $i$, we denote its feature at layer $r\in\mathcal{L}$ by $\mathbf{h}_i^{(r)}\in\mathbb{R}^{d_v}$, where $d_v$ is the VLM hidden dimension.
A joint router $\rho_\theta$, implemented as a two-layer MLP, takes the concatenated features of the same input token across candidate layers and produces an $N\times E$ matrix of layer--expert scores
\begin{equation}
 \mathbf{G}_i=\rho_\theta\!\left(
 [\mathbf{h}_i^{(r_1)};\ldots;\mathbf{h}_i^{(r_N)}]
 \right)\in\mathbb{R}^{N\times E},
 \label{eq:cae_joint_router}
\end{equation}
where $[\,;\,]$ denotes concatenation and $G_i^{(r,e)}$ scores expert $e$ at candidate layer $r$.
For each input token, we select the two highest-scoring experts in each candidate layer and denote the selected set by \(\mathcal{E}_i^{(r)}\).
We then use these expert scores to compute a layer score
\begin{equation}
 b_i^{(r)}=\log\!\sum_{e\in\mathcal{E}_i^{(r)}}
 \exp\!\left(G_i^{(r,e)}\right).
 \label{eq:cae_layer_score}
\end{equation}
We select the two layers with the highest $b_i^{(r)}$ and keep the two selected experts from each, forming the active set $\mathcal{S}_i$ of four layer--expert pairs.
For these four pairs, a single softmax over their original expert scores $G_i^{(r,e)}$ yields the fusion weights
\begin{equation}
 \alpha_i^{(r,e)}=
 \frac{\exp(G_i^{(r,e)})}
 {\sum_{(k,j)\in\mathcal{S}_i}\exp(G_i^{(k,j)})},
 \quad (r,e)\in\mathcal{S}_i.
 \label{eq:cae_fusion_weights}
\end{equation}
Each selected expert projects its own layer's feature, and the weighted outputs are fused as
\begin{equation}
 \mathbf{f}_i=\sum_{(r,e)\in\mathcal{S}_i}
 \alpha_i^{(r,e)}
 (\mathbf{W}_{r,e}\mathbf{h}_i^{(r)}+\mathbf{b}_{r,e}),
 \label{eq:cae_feature_fusion}
\end{equation}
where $\mathbf{W}_{r,e}\in\mathbb{R}^{d_f\times d_v}$ and $\mathbf{b}_{r,e}\in\mathbb{R}^{d_f}$ are the learnable weight and bias of expert $(r,e)$. The projected feature dimension $d_f$ matches the DiT hidden dimension.
The resulting features $\mathbf{F}_t=[\mathbf{f}_1,\ldots,\mathbf{f}_M]^{\mathsf T}\in\mathbb{R}^{M\times d_f}$ preserve the $M$ input-token positions, where $M$ is the input sequence length.
The routing configuration is provided in Sec.~\ref{supp:cae_configuration}.

\begin{figure*}[t]
  \centering
  \includegraphics[width=0.95\textwidth]{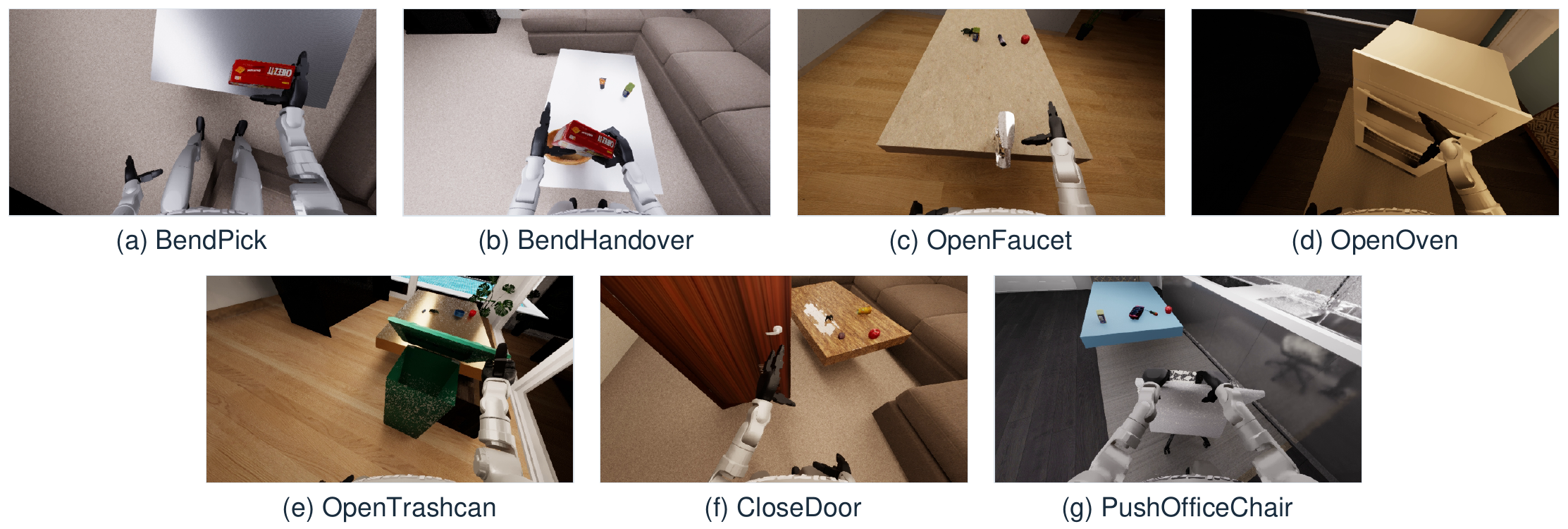}
  \caption{\textbf{Simulation task settings.} Rendered demonstration frames from SIMPLE illustrate the seven tasks.}
  \label{fig:simulation_tasks}
\end{figure*}

\begin{table*}[t]
\centering
\caption{Successful trials on seven SIMPLE tasks (30 per task, 10 per randomization level). Overall success is reported as successful/total trials and a percentage. Best and second-best results are bold and underlined, respectively.}
\label{tab:simulation_results}
\small
\setlength{\tabcolsep}{4pt}
\renewcommand{\arraystretch}{1.05}
\begin{tabular*}{\textwidth}{@{\extracolsep{\fill}}l*{9}{c}@{}}
\toprule
\multirow{2}{*}[-3pt]{Method} & \multirow{2}{*}[-3pt]{\shortstack{Bend\\Pick}} & \multirow{2}{*}[-3pt]{\shortstack{Bend\\Handover}} & \multirow{2}{*}[-3pt]{\shortstack{Open\\Faucet}} & \multirow{2}{*}[-3pt]{\shortstack{Open\\Oven}} & \multirow{2}{*}[-3pt]{\shortstack{Open\\Trashcan}} & \multirow{2}{*}[-3pt]{\shortstack{Close\\Door}} & \multirow{2}{*}[-3pt]{\shortstack{Push Office\\Chair}} & \multicolumn{2}{c}{Overall success} \\
\cmidrule(l){9-10}
 &  &  &  &  &  &  &  & Count & Rate (\%) \\
\midrule
$\pi_{0.5}$~\cite{black2025pi05} & 18/30 & 2/30 & \underline{22/30} & 18/30 & 8/30 & \underline{25/30} & \underline{27/30} & 120/210 & 57.1 \\
GR00T N1.6~\cite{nvidia2025grootn16} & \underline{24/30} & \underline{17/30} & 10/30 & 9/30 & \textbf{25/30} & 21/30 & 24/30 & \underline{130/210} & \underline{61.9} \\
$\Psi_0$~\cite{wei2026psi0} & 16/30 & 13/30 & 16/30 & \underline{19/30} & 20/30 & 17/30 & 22/30 & 123/210 & 58.6 \\
$\omega$-0~\cite{li2026omega0} & 22/30 & 3/30 & 12/30 & 9/30 & \underline{23/30} & 13/30 & 24/30 & 106/210 & 50.5 \\
DP~\cite{chi2023diffusionpolicy} & 0/30 & 16/30 & 2/30 & 10/30 & 19/30 & 3/30 & 20/30 & 70/210 & 33.3 \\
ACT~\cite{zhao2023act} & 19/30 & 8/30 & 15/30 & 18/30 & 20/30 & 24/30 & 18/30 & 122/210 & 58.1 \\
\midrule
CHAMP (ours) & \textbf{25/30} & \textbf{23/30} & \textbf{23/30} & \textbf{28/30} & 21/30 & \textbf{29/30} & \textbf{29/30} & \textbf{178/210} & \textbf{84.8} \\
\bottomrule
\end{tabular*}
\end{table*}

Conditioned on the fused features $\mathbf{F}_t$ and a token encoding the current robot state $\mathbf{s}_t$, the DiT generates the action chunk $\mathbf{A}_t$.
We train CAE using conditional flow matching~\cite{lipman2022flow}, which regresses a velocity field along a linear path between action chunks and Gaussian noise.
The overall training objective is
\begin{equation}
 \mathcal{L}_{\mathrm{ground}}
 =\mathcal{L}_{\mathrm{FM}}
  +\lambda_L\mathcal{L}_{\mathrm{bal}}^L
  +\lambda_E\mathcal{L}_{\mathrm{bal}}^E,
 \label{eq:action_grounding_loss}
\end{equation}
where $\mathcal{L}_{\mathrm{FM}}$ is the flow-matching loss,
the layer-level loss $\mathcal{L}_{\mathrm{bal}}^L$ discourages repeated selection of the same few candidate layers,
and the expert-level loss $\mathcal{L}_{\mathrm{bal}}^E$ discourages repeatedly choosing the same few projection experts within each selected layer.
The coefficients $\lambda_L$ and $\lambda_E$ weight the layer- and expert-level losses, respectively.
Detailed loss definitions and training settings are provided in the supplementary material (Secs.~\ref{supp:action_optimization} and~\ref{supp:flow_training}).

\subsection{Task Fine-tuning}
\label{sec:task_finetuning}

For each task, we fine-tune the policy learned during robot action grounding using simulation or real-world demonstrations.
Demonstrations are collected in both environments using a PICO headset and handheld controllers for whole-body teleoperation of a Unitree G1 equipped with Dex3 hands.
The demonstrations provide aligned RGB observations, instructions, robot states, and body and hand action sequences.
For the G1 robot with Dex3 hands, each action contains a 64-dimensional motion latent and seven joint commands per hand, giving $d_a=78$.
We keep the VLM frozen and fine-tune CAE with the objective in Eq.~\ref{eq:action_grounding_loss} to adapt both feature selection and action generation to each task.
Collection details are provided in Supplementary Sec.~\ref{supp:protocols}.

\section{Experiments}
\label{sec:experiments}

\subsection{Experimental Setup}
\label{sec:experimental_setup}
\label{sec:baselines}

\paragraph{Tasks and environments.}
We evaluate CHAMP in both simulation and real-world environments.
In simulation, we use seven humanoid loco-manipulation tasks from SIMPLE~\cite{wei2026simple}, illustrated in Fig.~\ref{fig:simulation_tasks}.
BendPick and BendHandover emphasize whole-body grasping and bimanual coordination, respectively.
OpenFaucet, OpenOven, and OpenTrashcan involve articulated-object manipulation.
CloseDoor and PushOfficeChair involve locomotion with object interaction.
Whole-body motion demonstrations for these tasks are collected through teleoperation.

\begin{figure*}[t]
  \centering
  \includegraphics[width=0.95\textwidth]{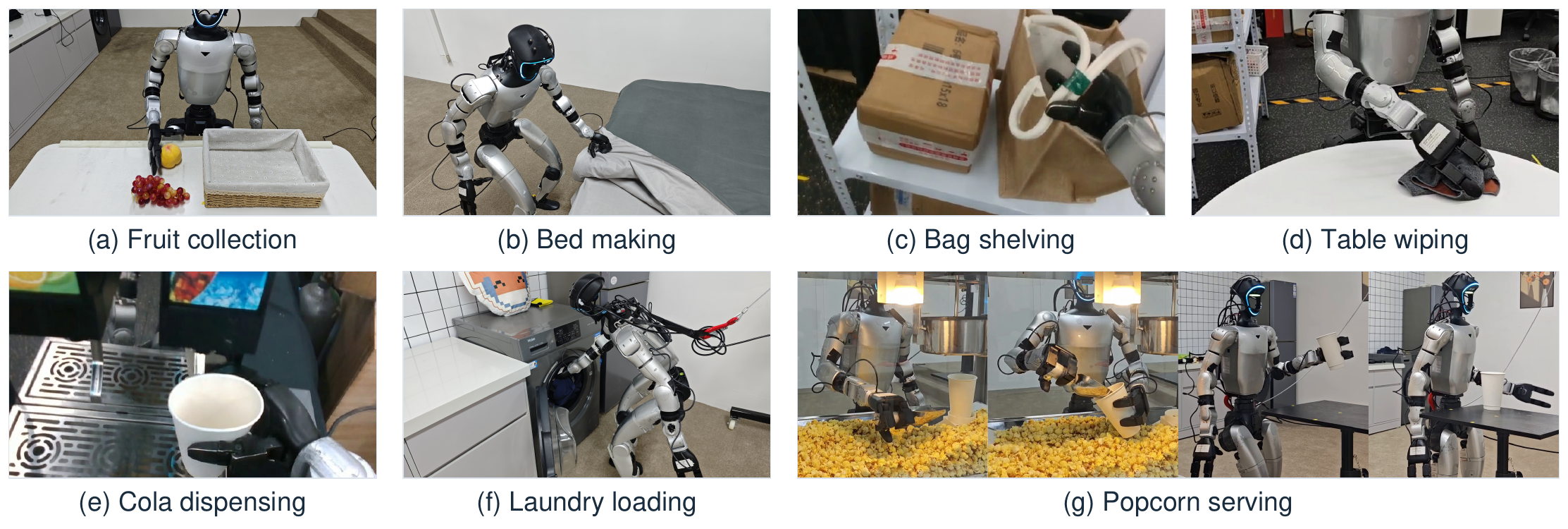}
  \caption{\textbf{Real-world task settings.} Images from our robot recordings illustrate the seven tasks. (g) shows successive stages of long-horizon popcorn serving.}
  \label{fig:real_world_tasks}
\end{figure*}

We evaluate seven real-world tasks on a Unitree G1 equipped with Dex3 hands, including the long-horizon popcorn serving task (Fig.~\ref{fig:real_world_tasks}).
The tasks span kitchens, bedrooms, living rooms, restaurants, laundry rooms, and industrial settings.
Fruit collection and bag shelving combine walking with object pickup and placement.
Bed making and table wiping involve coordinated torso and arm motion.
Cola dispensing involves positioning a cup under a dispenser, while laundry loading combines door operation and clothes handling with repeated adjustments of foot placement.
Detailed task procedures are provided in Sec.~\ref{supp:real_world_procedures} of the supplementary material.

\begin{table*}[t]
\centering
\caption{Successful trials on seven real-world tasks (10 per task). Overall success is reported as successful/total trials and a percentage. Best and second-best results are bold and underlined, respectively.}
\label{tab:real_world_results}
\small
\setlength{\tabcolsep}{4pt}
\renewcommand{\arraystretch}{1.05}
\begin{tabular*}{\textwidth}{@{\extracolsep{\fill}}l*{9}{c}@{}}
\toprule
\multirow{2}{*}[-3pt]{Method} & \multirow{2}{*}[-3pt]{\shortstack{Fruit\\collection}} & \multirow{2}{*}[-3pt]{\shortstack{Bed\\making}} & \multirow{2}{*}[-3pt]{\shortstack{Bag\\shelving}} & \multirow{2}{*}[-3pt]{\shortstack{Table\\wiping}} & \multirow{2}{*}[-3pt]{\shortstack{Cola\\dispensing}} & \multirow{2}{*}[-3pt]{\shortstack{Laundry\\loading}} & \multirow{2}{*}[-3pt]{\shortstack{Popcorn\\serving}} & \multicolumn{2}{c}{Overall success} \\
\cmidrule(l){9-10}
 &  &  &  &  &  &  &  & Count & Rate (\%) \\
\midrule
$\pi_{0.5}$~\cite{black2025pi05} & 5/10 & 6/10 & 4/10 & 6/10 & 6/10 & 4/10 & 2/10 & 33/70 & 47.1 \\
GR00T N1.6~\cite{nvidia2025grootn16} & \underline{7/10} & \underline{9/10} & \underline{8/10} & \underline{9/10} & \underline{9/10} & \underline{6/10} & \underline{4/10} & \underline{52/70} & \underline{74.3} \\
$\Psi_0$~\cite{wei2026psi0} & 3/10 & 7/10 & 6/10 & \underline{9/10} & 7/10 & 5/10 & 3/10 & 40/70 & 57.1 \\
$\omega$-0~\cite{li2026omega0} & 5/10 & 7/10 & 7/10 & 8/10 & 8/10 & 5/10 & 3/10 & 43/70 & 61.4 \\
DP~\cite{chi2023diffusionpolicy} & 0/10 & 2/10 & 6/10 & 0/10 & 3/10 & 0/10 & 0/10 & 11/70 & 15.7 \\
ACT~\cite{zhao2023act} & 4/10 & 7/10 & \underline{8/10} & 6/10 & 1/10 & 4/10 & 1/10 & 31/70 & 44.3 \\
\midrule
CHAMP (ours) & \textbf{9/10} & \textbf{10/10} & \textbf{9/10} & \textbf{10/10} & \textbf{10/10} & \textbf{9/10} & \textbf{7/10} & \textbf{64/70} & \textbf{91.4} \\
\bottomrule
\end{tabular*}
\end{table*}

\paragraph{Baselines and evaluation.}
We compare CHAMP with six baselines.
$\pi_{0.5}$~\cite{black2025pi05} and GR00T N1.6~\cite{nvidia2025grootn16} are generalist vision-language-action models.
$\Psi_0$~\cite{wei2026psi0} and $\omega$-0~\cite{li2026omega0} target humanoid loco-manipulation, while Diffusion Policy (DP)~\cite{chi2023diffusionpolicy} and Action Chunking with Transformers (ACT)~\cite{zhao2023act} are conventional imitation-learning methods.
We train all baselines using their official implementations, adapted to the target robot's control interface.
For each task, all methods use the same 100 demonstrations and number of training steps.

We run 30 trials per simulation task, with 10 at each of three environment randomization levels, and 10 trials per real-world task.
Overall success is the mean of per-task success rates across seven tasks in each setting.

\subsection{Simulation Results}
\label{sec:simulation}
CHAMP achieves the highest overall success rate of 84.8\% and ranks first on six of the seven simulation tasks (Table~\ref{tab:simulation_results}).
It outperforms the best-performing baseline, GR00T N1.6 (61.9\%), by 22.9 percentage points.
CHAMP also outperforms the humanoid loco-manipulation baselines $\Psi_0$ (58.6\%) and $\omega$-0 (50.5\%).
On OpenOven, which requires body adjustments during object manipulation, CHAMP succeeds in 28 of 30 trials compared with 19 for $\Psi_0$, yielding the largest gain over the best baseline across tasks.
On bimanual BendHandover, CHAMP succeeds in 23 of 30 trials, compared with 17 for GR00T N1.6.
It also succeeds in 29 of 30 trials on each of CloseDoor and PushOfficeChair, which combine locomotion with object interaction.
The only exception is OpenTrashcan, where CHAMP succeeds in 21 of 30 trials compared with 25 for GR00T N1.6.

\subsection{Real-World and Long-Horizon Results}
\label{sec:real_world}
\label{sec:long_horizon}

CHAMP achieves a mean success rate of 91.4\% across the seven real-world tasks in Table~\ref{tab:real_world_results}, completing 64 of 70 trials.
It outperforms GR00T N1.6 (74.3\%) by 17.1 percentage points and ranks first on every task.
CHAMP also outperforms the humanoid loco-manipulation baselines $\Psi_0$ and $\omega$-0, which achieve mean success rates of 57.1\% and 61.4\%, respectively.

On laundry loading, CHAMP succeeds in 9 of 10 trials compared with 6 for GR00T N1.6.
During laundry loading, the robot must repeatedly adjust its foot placement to open the washing machine door, move closer to load clothes, and step backward while closing the door, requiring flexible lower-body control.
During fruit collection, which requires walking to the table and then grasping the fruits, CHAMP succeeds in 9 of 10 trials compared with 7 for GR00T N1.6.
These gains span different coordination demands, including body adjustment in bed making and walking while carrying objects in bag shelving.

Interestingly, we observed adaptive responses that were absent from the task demonstrations.
When an experimenter wiped up the cola spill as the robot reached for a cleaning cloth, the robot withdrew its hand instead of proceeding with wiping.
During fruit collection, if the robot failed to grasp the fruit, it would autonomously try several more times to complete the task.
These qualitative observations suggest that CHAMP can adjust its ongoing behavior in response to scene changes and failed interactions.

\paragraph{Long-horizon task.}
Popcorn serving requires the robot to grasp a scoop, pick up a bucket with its other hand, scoop popcorn, and pour it into the bucket (Fig.~\ref{fig:real_world_tasks}g).
It then walks to a table while carrying the filled bucket and sets it down.
This sequence involves multiple object interactions and bimanual coordination.
The robot must also place the bucket securely on the table to avoid spilling the popcorn.
A trial is successful only if the robot completes the entire sequence.
CHAMP completes the full task in 7 of 10 trials, compared with 4 for GR00T N1.6 and 3 each for $\Psi_0$ and $\omega$-0.
The conventional imitation-learning baselines DP and ACT succeed in 0 and 1 of 10 trials, respectively.

In one of CHAMP's three failed trials, an unsuitable scoop grip prevented scooping popcorn.
In the other two, excessive scoop–bucket contact force during pouring caused the robot to lose its grip.
Such interactions are difficult to control, and similar failures can occur even during teleoperation because no force feedback is available.

\subsection{Ablation Studies}
\label{sec:ablations}
To assess the contributions of whole-body motion pretraining, robot action grounding, and CAE's multi-layer feature extraction, we conduct ablation experiments on the seven SIMPLE tasks (Table~\ref{tab:simulation_ablations}).
The variant without motion pretraining initializes the VLM from the original Qwen3-VL checkpoint and retains robot action grounding and task fine-tuning.
Its overall success rate falls from 84.8\% to 71.9\%, indicating that body and hand motion pretraining provides useful knowledge for downstream control.

The variant without robot action grounding retains the motion-pretrained VLM but skips CAE training on replay data.
We fine-tune the policy directly on teleoperated task demonstrations.
This variant achieves an overall success rate of 73.8\%.
This result indicates that robot action grounding helps apply the learned motion priors to robot control and improves task success rates after fine-tuning.

We then evaluate CAE's feature extraction using four fixed-layer variants.
All four variants retain motion pretraining and robot action grounding, but replace CAE with a standard action DiT conditioned on features from a fixed VLM layer (4, 12, 20, or 28).
They achieve success rates of 62.4--78.6\%, with layer 4 performing best among the tested layers.
The full model outperforms the best fixed-layer variant (layer 4) and the final-layer variant (layer 28) by 6.2 and 22.4 percentage points, respectively.
These results show that CAE’s adaptive multi-layer feature extraction achieves higher task success rates than the fixed-layer variants.
Further routing analyses appear in Supplementary Sec.~\ref{supp:routing_analysis}.
\begin{table}[t]
\centering
\caption{Ablations on seven SIMPLE tasks (210 trials per variant). Decrease denotes the success-rate drop from full CHAMP in percentage points, computed from trial counts.}
\label{tab:simulation_ablations}
\small
\setlength{\tabcolsep}{3pt}
\renewcommand{\arraystretch}{1.05}
\begin{tabular*}{\columnwidth}{@{\extracolsep{\fill}}lcc@{}}
\toprule
Variant & Success (\%) & Decrease \\
\midrule
w/o motion pretraining & 71.9 & 12.9 \\
w/o robot action grounding & 73.8 & 11.0 \\
\midrule
Fixed layer 4 & 78.6 & 6.2 \\
Fixed layer 12 & 74.3 & 10.5 \\
Fixed layer 20 & 65.2 & 19.5 \\
Fixed layer 28 & 62.4 & 22.4 \\
\midrule
CHAMP (full) & \textbf{84.8} & --- \\
\bottomrule
\end{tabular*}
\end{table}

\section{Conclusion}
\label{sec:conclusion}
To enable coordinated locomotion and dexterous manipulation, we presented CHAMP, a unified framework for humanoid loco-manipulation.
CHAMP pretrains a VLM on heterogeneous human and robot data to learn whole-body motion priors through body and hand motion prediction.
We introduce CAE and train it on replayed robot data to ground the learned motion priors in robot control.
This action expert adaptively selects, transforms, and fuses features from the motion-pretrained VLM to generate coordinated body and hand actions.
CHAMP achieves overall success rates of 84.8\% on seven simulated tasks and 91.4\% on seven real-world tasks, including 70.0\% on long-horizon popcorn serving.
Ablation studies further demonstrate the benefits of motion pretraining and robot action grounding, as well as the advantage of CAE over fixed-layer feature extraction.
Together, these results highlight the value of combining whole-body motion learning with adaptive feature extraction for coordinated humanoid loco-manipulation.

\paragraph{Limitations and future work.}
Our motion pretraining focuses on body and hand kinematics without explicit supervision of object dynamics or contact forces.
Control evaluation is also limited to a single humanoid platform with task-specific fine-tuning.
Future work will explore object- and contact-aware motion priors and generalization across embodiments and unseen tasks.

\section*{LLM Usage Statement}
We worked closely with OpenAI's ChatGPT and Codex to draft and iteratively refine the writing throughout the manuscript, improving clarity, precision, and coherence. These tools also assisted with figure preparation and code development for data analysis and visualization. The authors take full responsibility for the accuracy and integrity of the final manuscript.

{
    \small
    \bibliographystyle{ieeenat_fullname}
    \bibliography{references}
}

\providecommand{\suppadd}[1]{#1}
\renewcommand{\suppadd}[1]{#1}
\clearpage
\maketitlesupplementary
\appendix

\section{Data Preparation and Supervision}
\label{supp:data}
\subsection{VLM Pretraining Data and Pose Representation}
\label{supp:motion_data}

\paragraph{Data and temporal alignment.}
The first-stage VLM is pretrained to predict future motion from an RGB image, a language instruction, and pose inputs. We first convert the source recordings into a shared pose representation, from which the VLM inputs and prediction targets are constructed. The nine sources in Table~\ref{tab:supp_motion_sources} contain 2,297.9 hours across 20 processed subsets. All trajectories are stored at 30\,Hz; resampling uses linear interpolation for positions and spherical interpolation for rotations.

\begin{table*}[t]
\centering
\caption{Pretraining data sources and motion annotations.
Durations are rounded independently.
UnifoLM-WBT contributes 6.37 hours with BrainCo hands, 7.11 hours with Inspire hands, and 1.14 hours with Dex1 grippers.
The first two subsets provide body and hand targets, while Dex1 provides body targets only.
The simulated GR1 data comprise the arms-only, arms-and-waist, and unified subsets.}
\label{tab:supp_motion_sources}
\small
\setlength{\tabcolsep}{7pt}
\begin{tabular}{llrl}
\toprule
Source & Domain / embodiment & Hours & Available targets \\
\midrule
EgoDex & Human & 655.9 & Hand \\
Xperience-10M & Human & 326.3 & Body and hand \\
ARCTIC & Human & 1.8 & Body and hand \\
HIW-500 & Unitree G1 with Dex1 grippers & 335.4 & Body \\
HumanoidEveryday & Unitree G1 with Dex3 / H1 with Inspire & 31.5 & Body and hand \\
UnifoLM-WBT & Unitree G1 with BrainCo / Dex1 / Inspire & 14.6 & Depends on end effector \\
PSI-Real & Unitree G1 with Dex3 & 7.7 & Hand \\
RoboCOIN & Selected Unitree G1 with Dex3 recordings & 2.5 & Hand \\
GR00T-X-Embodiment-Sim & Simulated Fourier GR1 & 922.1 & Body and hand \\
\midrule
Total & & 2,297.9 & \\
\bottomrule
\end{tabular}

\end{table*}

\paragraph{Pose representation for motion supervision.}
To construct the VLM's motion targets, we represent pelvis and wrist poses relative to the camera pose at the start of each episode. This fixed reference frame is denoted by $G=C_0$ and remains unchanged as the camera moves. Its axes follow the OpenCV convention: $+x$ right, $+y$ down, and $+z$ forward. Positions are measured in meters. A source pelvis or wrist pose at time $u$ is converted to this reference frame as
\begin{equation}
 {}^{G}\!T_{S_u}=({}^{W}\!T_{C_0})^{-1}{}^{W}\!T_{S_u},
 \label{eq:supp_reference_frame}
\end{equation}
where $W$ is the source dataset's world frame, $S_u$ is the pelvis or wrist frame, and ${}^{A}\!T_B$ maps coordinates from $B$ to $A$. We denote the resulting position and canonical orientation by $\mathbf r_u^G$ and $R_u$. Thus, $W$ is used only to convert source poses into the fixed camera reference.

Local articulation is represented separately: body-point coordinates are relative to the pelvis at the same timestamp, and fingertip coordinates are relative to the corresponding wrist at that timestamp. These coordinates, denoted by $\mathbf b_u$, describe body and finger configurations relative to their parent joints.

\suppadd{We standardize these joint-attached axes across sources. The pelvis frame uses x forward, y left, and z up. In each wrist frame, x points distally toward the fingers and z radially toward the thumb; y completes the right-handed basis as z cross x. These directions define the joint-attached frames, rather than the fixed camera reference. A source-specific constant rotation converts each native joint basis to this convention. We right-multiply the native orientation by this rotation and apply its inverse to the local point coordinates, leaving the reconstructed positions in the fixed camera frame unchanged.}

The resulting body vector contains 45 values: pelvis position (3), 6D rotation (6), and 12 local body points (36). The points are ordered as the left shoulder, elbow, wrist, hip, knee, and ankle, followed by the corresponding right-side joints. The hand vector contains 48 values, with the left hand before the right. Each hand contributes wrist position (3), 6D rotation (6), and five local fingertip positions (15), ordered thumb, index, middle, ring, and little.

\paragraph{Source-specific processing.}
EgoDex provides ARKit-world wrist and fingertip transforms.
ARCTIC uses personalized SMPL-X and MANO models to recover body joints, wrists, and fingertip vertices.
For Xperience-10M, captions define action intervals, while SLAM poses and calibrated stereo-left camera-to-device transforms determine camera poses. SMPL-H roots and keypoints provide body geometry; wrists and metacarpals define palm frames, and valid MANO annotations provide fingertip positions.

HIW-500 uses matched G1/Dex1 forward kinematics from 29 motor positions and odometry pelvis poses, synchronized to head-image timestamps with a nominal D435 camera mount.
HumanoidEveryday uses matched G1/Dex3 and H1/Inspire models with recorded arm, leg, and hand states.
PSI-Real combines arm forward kinematics with recorded torso orientation and height.
RoboCOIN aligns source end-effector poses with bilateral G1/Dex3 wrist forward kinematics before applying the nominal D435 mount.
UnifoLM-WBT uses floating-base poses and 29 body joints, with matched BrainCo or Inspire hand models and the wrist-yaw link as wrist root.
The three GR00T-X-Embodiment-Sim subsets use Fourier GR1 forward kinematics with complete leg coordinates.

\subsection{VLM Pretraining Inputs and Motion Targets}
\label{supp:motion_tokenization}
Each VLM pretraining sample pairs an image, an instruction, and pose inputs at the observation time with a sequence of future motion targets. We construct the targets from the pose representation above and encode them with FAST. The pose inputs describe wrist poses relative to the current camera and the recent relative motion of the pelvis.

\paragraph{Relative motion targets.}
Motion prediction uses a 5\,Hz grid, with prediction index $t$ anchored at frame $n=6t$ of the processed 30\,Hz trajectory. The five targets at $n+6,n+12,n+18,n+24,n+30$ span the following second at 0.2-second intervals, corresponding to $t+1,\ldots,t+5$ in Eq.~\ref{eq:available_motion_targets}. The anchor frame provides the observation and reference pose; windows with incomplete future horizons are excluded.

For $k\in\{1,\ldots,5\}$, pelvis and wrist translation targets are $\mathbf r_{t+k}^G-\mathbf r_t^G$, expressed in the fixed reference frame $G$. Local-point targets are $\mathbf b_{t+k}-\mathbf b_t$, with each term expressed in the canonical pelvis or wrist frame at the corresponding timestamp. These differences capture changes in local articulation. Relative orientations are represented by
\begin{equation}
 R^{\mathrm{rel}}_{t,k}=R_t^{\mathsf T}R_{t+k},
 \qquad
 \operatorname{r6}(R)=
 \begin{bmatrix}R_{:,1}\\R_{:,2}\end{bmatrix}.
 \label{eq:supp_relative_rotation}
\end{equation}
The first two columns of each relative rotation matrix are concatenated to form the six-dimensional representation used for normalization and FAST encoding.

\paragraph{Pose inputs to the VLM.}
Alongside the RGB image and instruction, the VLM receives the available wrist and pelvis features. Wrist inputs describe hand position and orientation relative to the camera at observation time $t$. We denote this current camera frame by $C_t$; unlike the fixed first-camera frame $G$, it follows the camera as it moves. For the canonical wrist frame $S_t^h$ of hand $h\in\{\mathrm L,\mathrm R\}$, the input pose is
\begin{equation}
 {}^{C_t}\!T_{S_t^h}=({}^{G}\!T_{C_t})^{-1}{}^{G}\!T_{S_t^h}.
 \label{eq:supp_prompt_wrist_pose}
\end{equation}
The pelvis input represents the current pelvis pose relative to the pelvis frame 0.2s earlier. Let $P_t$ denote the canonical pelvis frame at prediction index $t$. This input is the pose of $P_t$ relative to $P_{t-1}$, the pelvis frame at the preceding point on the 5\,Hz grid:
\begin{equation}
 D_t=({}^{G}\!T_{P_{t-1}})^{-1}{}^{G}\!T_{P_t},\qquad t\geq1,
 \label{eq:supp_prompt_pelvis_delta}
\end{equation}
with $D_0$ set to the identity transform at the first prediction anchor. Each transform is represented by three translation coordinates and six rotation coordinates using Eq.~\ref{eq:supp_relative_rotation}, with the left wrist preceding the right.

\paragraph{Normalization and FAST encoding.}
We fit separate body and hand tokenizers, each with its own normalization statistics and vocabulary, and reserve 10\% of episodes for validation. For each modality, we sample 1\% of eligible training windows in proportion to the candidate count of each source, yielding 1,732,393 hand windows and 1,482,427 body windows. A relative target value $v_d$ in dimension $d$ is normalized using its 1st and 99th percentiles, $q_{0.01,d}$ and $q_{0.99,d}$:
\begin{equation}
 \widetilde v_d=\operatorname{clip}\!\left(
 \frac{v_d-\tfrac12(q_{0.01,d}+q_{0.99,d})}
 {\tfrac12(q_{0.99,d}-q_{0.01,d})},-1,1\right).
 \label{eq:supp_motion_normalization}
\end{equation}
FAST applies an orthonormal temporal DCT to each normalized dimension, scales the coefficients by 10, rounds them to integers, and flattens them in time-major order. The resulting sequence is tokenized using byte-pair encoding with a vocabulary of 8,192 tokens per modality.

\subsection{SONIC Motion Encoding, Replay, and Data Filtering}
\label{supp:sonic_replay}
The second stage trains CAE on robot control targets while keeping the pretrained VLM frozen. For this action-grounding stage, we use HIW-500, HumanoidEveryday-G1, and UnifoLM, including the original UnifoLM-WBT subsets and additional G1 recordings with Inspire hands or Dex1 grippers.

\paragraph{Motion encoding.}
We encode recorded body motion using SONIC's low-latency G1 encoder~\cite{luo2025sonic}. For each frame $n$ on the 30\,Hz trajectory, the encoder receives a ten-frame window beginning at $n$, containing 29 body-joint positions in the G1 joint order, finite-difference joint velocities, and root orientations. Root rotations are expressed relative to the first frame of the window:
\begin{equation}
 \Delta R_{n,k}=R_n^{\mathsf T}R_{n+k},
 \qquad k\in\{0,\ldots,9\},
 \label{eq:supp_sonic_relative_rotation}
\end{equation}
where $R_n$ denotes the reference root rotation in the SONIC coordinate convention. The first two columns of each relative rotation are flattened in row-major order to form a six-dimensional input. Windows extending beyond the segment boundary repeat the final frame. The encoder produces one 64-dimensional latent $\mathbf z_n$ per frame for body-action supervision.

Each latent is paired with source-derived end-effector targets: one channel per hand for grippers, six for BrainCo and Inspire hands, and seven Dex3 joint-angle channels for HumanoidEveryday-G1. The resulting action dimensions are $64+2=66$, $64+12=76$, and $64+14=78$. State vectors concatenate 29 body-joint angles with the corresponding end-effector states, giving dimensions of 31, 41, and 43. These heterogeneous interfaces are accommodated by the shared model described in Sec.~\ref{supp:action_optimization}.

\paragraph{Closed-loop screening.}
We screen segments for instability and excessive root displacement through closed-loop replay in MuJoCo. A G1 initialized from the first reference body pose is driven by the precomputed latents through the SONIC controller at 50\,Hz, with latents selected from the 30\,Hz reference sequence by time index. Closed-loop replay determines which segments are retained; the corresponding source data are then used for training.

Let $d_{\mathrm{ref}}$ and $d_{\mathrm{sim}}$ be the magnitudes of the net horizontal root displacement in the reference and simulated trajectories. A segment is rejected if the simulated root height falls below $0.35$\,m, if $d_{\mathrm{ref}}<0.15$\,m and $d_{\mathrm{sim}}>0.25$\,m, or if $d_{\mathrm{sim}}-d_{\mathrm{ref}}>0.5$\,m.

The resulting grounding corpus contains 40,016 of 60,686 candidate segments across seven subsets, comprising 27,482,899 frames at 30\,Hz, or 254.47 hours (Table~\ref{tab:supp_action_grounding_inventory}). Task-specific fine-tuning demonstrations are separate from this corpus.

\begin{table}[t]
\centering
\caption{Action-grounding corpus after episode-level screening. Durations are computed from retained frame counts and rounded independently. Actions comprise a 64-dimensional SONIC latent and source-specific end-effector channels.}
\label{tab:supp_action_grounding_inventory}
\small
\setlength{\tabcolsep}{2.25pt}
\begin{tabular}{lrrr}
\toprule
Source subset & Segments & Hours & Action dim.\\
\midrule
HIW-500 & 8,119 & 93.27 & 66\\
HumanoidEveryday-G1 & 1,514 & 5.41 & 78\\
UnifoLM-WBT / BrainCo & 612 & 4.68 & 76\\
UnifoLM-WBT / Inspire & 519 & 4.15 & 76\\
UnifoLM-WBT / Dex1 & 23 & 0.18 & 66\\
UnifoLM-G1 / Inspire & 168 & 1.97 & 76\\
UnifoLM-G1 / Dex1 & 29,061 & 144.81 & 66\\
\midrule
Total & 40,016 & 254.47 & --\\
\bottomrule
\end{tabular}

\end{table}

\section{Implementation Details}
\label{supp:implementation}
\subsection{Model Architecture and Routing}
\label{supp:cae_configuration}
\paragraph{Backbone and feature readout.}
We use Qwen3-VL-2B-Instruct and extract the outputs of decoder blocks $\mathcal L=\{4,12,20,28\}$. Each output, including that of block 28, passes once through the backbone's frozen final normalization. The normalized feature is shared by the router and the projection experts assigned to that layer.

\paragraph{Routing and action generation.}
CAE connects the frozen VLM to the action DiT through sparse cross-layer projection (Table~\ref{tab:supp_cae_configuration}). Each expert is an independent affine map from one candidate layer to the action-conditioning space. This structure extends single-layer MoE projection, in which all experts receive features from the same depth, and layer-only fusion, which assigns a single projection to each layer.

A joint two-layer MLP with SiLU activation concatenates features at each token position across the candidate layers and scores all layer--expert pairs. Within each layer, it selects two experts and computes a layer score as the log-sum-exp of their scores. The two highest-scoring layers define four active pairs, whose scores are jointly normalized by a single softmax at temperature $1$ (Eqs.~\ref{eq:cae_layer_score}--\ref{eq:cae_feature_fusion}). Only the selected experts are evaluated for each token. Their weighted sum directly produces a $1536$-dimensional conditioning feature, preserving token order and sequence length.

A linearly projected robot-state token is appended to the fused conditioning sequence. Noisy actions are embedded by an MLP with learned temporal position embeddings. Each DiT block applies noncausal joint attention to action and conditioning tokens, with separate query/key/value projections for the two streams and masking of padded conditioning tokens. Flow time modulates adaptive normalization.

\begin{table}[t]
\centering
\caption{CAE architecture and G1--Dex3 interfaces. State dimensions correspond to measured joint angles; action dimensions comprise the SONIC latent and bilateral hand targets.}
\label{tab:supp_cae_configuration}
\small
\setlength{\tabcolsep}{3pt}
\begin{tabular}{@{}ll@{}}
\toprule
Component & Configuration \\
\midrule
VLM hidden dimension & $d_v=2048$ \\
Candidate layers / experts per layer & $N=4$ / $E=8$ \\
Router dimensions & $8192\rightarrow128\rightarrow32$ \\
Each affine projection expert & $2048\rightarrow1536$ \\
Active layers / experts per layer & $K_L=2$ / $K_E=2$ \\
DiT blocks / hidden dimension & $6$ / $d_f=1536$ \\
Attention heads / head dimension & $24$ / $64$ \\
Action horizon & $T_a=30$ \\
G1--Dex3 state dimension & $43$ \\
G1--Dex3 action dimension & $d_a=64+7+7=78$ \\
\bottomrule
\end{tabular}

\end{table}

\subsection{Training and Fine-tuning Configurations}
\label{supp:action_optimization}
\paragraph{Optimization.}
Table~\ref{tab:supp_training_configuration} summarizes the settings for first-stage VLM pretraining, second-stage action grounding, and task-specific fine-tuning. VLM pretraining updates all VLM parameters while keeping the fitted FAST tokenizers fixed, using bfloat16 mixed precision, gradient checkpointing, and DeepSpeed ZeRO-3. We clip the gradient norm to $1.0$ and use a maximum sequence length of $8192$ tokens.

Robot action grounding and simulation fine-tuning optimize CAE with Eq.~\ref{eq:action_grounding_loss} while keeping the VLM frozen. Both use AdamW with $\epsilon=10^{-8}$, bfloat16 mixed precision, and gradient-norm clipping at $1.0$. For each simulation task, CAE is initialized from the 100,000-step grounding checkpoint and the optimizer is reset. Router scores, probabilities, and balancing statistics are computed in float32.

\paragraph{\suppadd{Real-world fine-tuning.}}
\suppadd{Real-world task policies initialize CAE from the action-grounding checkpoint and retain the frozen motion-pretrained VLM. We optimize CAE using eight GPUs with 16 samples per GPU, yielding a global batch size of 128. AdamW uses coefficients $(0.95,0.999)$, weight decay $10^{-6}$, and $\epsilon=10^{-8}$. The learning rate increases during warmup to $10^{-4}$ and then follows cosine decay. Training uses bfloat16 precision and gradient-norm clipping at 1.0. All seven tasks retain the 43-dimensional state, 78-dimensional action, and 30-frame action horizon.}

\paragraph{Image preprocessing.}
For VLM pretraining and action grounding, images are resized with Lanczos interpolation according to their source resolution (Table~\ref{tab:supp_image_resize}), followed by patch-aligned resizing with the Qwen image processor within a pixel range of $1{,}024$--$129{,}600$.

\begin{table}[t]
\centering
\caption{Image resizing before the Qwen image processor. Motion pretraining and robot action grounding use the same resolution-dependent policy.}
\label{tab:supp_image_resize}
\small
\begin{tabular}{@{}ll@{}}
\toprule
Source resolution & Resized resolution \\
\midrule
$1920\times1080$ & $480\times270$ \\
$840\times600$ & $420\times300$ \\
$640\times480$ & $320\times240$ \\
$512\times512$ & $256\times256$ \\
$256\times256$ & $256\times256$ \\
\bottomrule
\end{tabular}

\end{table}
Simulation frames are decoded at their aligned observation indices, downsampled from $640\times360$ to $320\times180$ with Lanczos interpolation, and cached as RGB before augmentation. Training reads the shared cache and applies online color jitter, with brightness, contrast, and saturation factors in $[0.8,1.2]$ and hue shifts in $[-0.05,0.05]$.

\begin{table*}[t]
\centering
\caption{Optimization settings for motion pretraining, action grounding, and per-task simulation fine-tuning. Fine-tuning uses a batch size of $16$ per GPU without gradient accumulation. The pretraining and grounding columns describe the checkpoints used to initialize the subsequent stage.}
\label{tab:supp_training_configuration}
\small
\setlength{\tabcolsep}{6pt}
\begin{tabular}{llll}
\toprule
Setting & Motion pretraining & Robot action grounding & Simulation fine-tuning \\
\midrule
Trainable parameters & Full VLM & CAE & CAE \\
AdamW $(\beta_1,\beta_2)$ & $(0.9,0.999)$ & $(0.9,0.999)$ & $(0.95,0.999)$ \\
Weight decay & $0.01$ & $0$ & $10^{-6}$ \\
Language-model learning rate & $10^{-4}$ & Frozen & Frozen \\
Visual-module learning rate & $10^{-5}$ & Frozen & Frozen \\
CAE learning rate & -- & $10^{-4}$ & $10^{-4}$ \\
Learning-rate schedule & Constant & Constant & Warmup + cosine decay \\
Warmup steps & $0$ & $0$ & $1000$ \\
Global batch size & $768$ & $640$ & $128$ \\
Training steps & $250{,}000$ & $100{,}000$ & $20{,}000$ \\
Training GPUs & $48\times$ NVIDIA H20 & $40\times$ NVIDIA H20 & $8$ GPUs \\
Training time & $150$ hours & $40$ hours & $2$ hours \\
\bottomrule
\end{tabular}

\end{table*}

\paragraph{Heterogeneous action and state interfaces.}
The grounding datasets provide native action dimensions of $66$, $76$, or $78$ and state dimensions of $31$, $41$, or $43$ (Sec.~\ref{supp:sonic_replay}). We accommodate all sources with a shared $78$-dimensional action head and $43$-dimensional state interface. Native vectors occupy the leading channels, with trailing channels zero-padded. Actions retain the source order of the 64-dimensional SONIC latent followed by left- and right-end-effector targets; states retain the 29 body-joint entries followed by the corresponding end-effector states.

\subsection{Flow-Matching and Load-Balancing Losses}
\label{supp:flow_training}
\paragraph{Action normalization.}
Grounding actions are normalized independently for each source and dimension to $[-1,1]$ using min--max bounds, with out-of-range values clipped. We denote a normalized, padded action chunk by $\bar{\mathbf A}_t\in\mathbb R^{T_a\times d_a}$, with $d_a=78$.

\suppadd{During grounding, the numerical joint-state vector is zero-padded to the shared state dimension and passed to the state projection without an additional min--max transform. This numerical input is distinct from the pose text supplied to the VLM.}

\paragraph{Flow-matching objective.}
For each grounding chunk, we sample independent standard-normal noise $\boldsymbol\epsilon\in\mathbb R^{T_a\times d_a}$ and a flow time

\begin{equation}
 \sigma=0.01+0.99\,\xi,\qquad \xi\sim\operatorname{Beta}(1.5,1).
 \label{eq:supp_flow_time}
\end{equation}
The sampled flow time is shared by all action positions. The linear interpolation and target velocity are

\begin{equation}
 \begin{aligned}
  \mathbf X_t^\sigma&=(1-\sigma)\bar{\mathbf A}_t+\sigma\boldsymbol\epsilon,\\
  \mathbf V_t&=\boldsymbol\epsilon-\bar{\mathbf A}_t.
 \end{aligned}
 \label{eq:action_flow_path}
\end{equation}
The DiT predicts $\widehat{\mathbf V}_t=v_\theta(\mathbf X_t^\sigma,\sigma;\mathbf F_t,\mathbf s_t)$. We minimize

\begin{equation}
 \mathcal L_{\mathrm{FM}}=
 \mathbb E\!\left[\frac{1}{d_a}
 \sum_{j=0}^{T_a-1}\sum_{c=1}^{d_a}m_{t,j,c}
 \bigl(\widehat V_{t,j,c}-V_{t,j,c}\bigr)^2\right],
 \label{eq:action_flow_loss}
\end{equation}
where $m_{t,j,c}$ is a binary validity mask.
\suppadd{For grounding, the mask is one for native action channels and zero for trailing padded channels at every position. We sample only complete action windows, so no temporal padding is needed. The loss sums over action positions and averages over samples and the shared action dimension, rather than dividing by the number of valid channels.}

\paragraph{Simulation fine-tuning.}
For each task, we recompute per-dimension min--max bounds from its demonstrations and normalize both the $43$-dimensional state and $78$-dimensional action to $[-1,1]$, clipping values to this interval. Fine-tuning samples $\sigma\sim\mathcal U(0,1)$ and a clean-prefix length $d$ uniformly from $\{0,\ldots,7\}$. The first $d$ actions remain uncorrupted, while the remaining positions share the sampled flow time.

\paragraph{\suppadd{Real-world fine-tuning.}}
\suppadd{For real-world tasks, actions are normalized using task-specific per-dimension min--max bounds and clipped to $[-1,1]$. The six tasks other than popcorn serving retain unscaled numerical joint states, as in grounding; popcorn serving instead applies task-specific min--max normalization and clipping to its state vector. Flow time follows the same rescaled Beta distribution as grounding, with a lower bound of 0.01. We also use a clean prefix of uniformly sampled length from 0 to 7 frames: prefix positions receive zero flow time and remain uncorrupted, while the remaining positions share the sampled flow time. Clean-prefix positions and action positions extending beyond the episode boundary are excluded from the flow-matching loss. The layer- and expert-balancing coefficients remain unchanged.}

\paragraph{Routing statistics.}
Load balancing is computed over the set $\mathcal I$ of all valid VLM token occurrences across samples and devices in the current forward pass, with $n=|\mathcal I|$. Each token selects $K_L=2$ layers and $K_E=2$ experts per selected layer. Let $\delta_{ir}$ indicate whether layer $r$ belongs to token $i$'s final active set $\mathcal S_i$. The layer count is $m_r=\sum_{i\in\mathcal I}\delta_{ir}$, and $c_{r,e}$ counts tokens for which $(r,e)\in\mathcal S_i$.

\paragraph{Layer-level balancing.}
\begin{figure*}[!t]
\centering
\includegraphics[width=\textwidth]{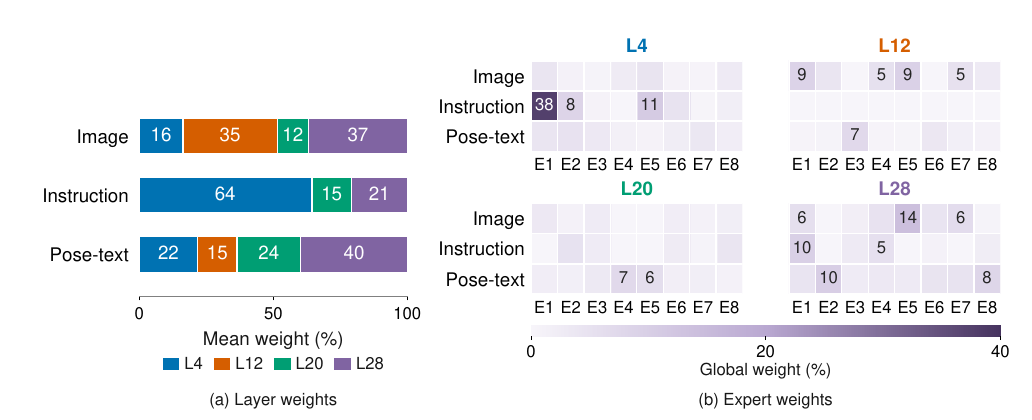}
\caption{\textbf{Layer and expert fusion weights.} Weights are averaged separately by token category and equally across the seven task models. (a) Layer weights sum to 100\% within each category. (b) Expert weights sum to 100\% across all 32 experts within each category and use a shared color scale; E1--E8 identify experts within each layer. Values are rounded percentages, with labels omitted below 5\%.}
\label{fig:supp_cae_routing_profiles}
\end{figure*}

The layer scores $b_i^{(r)}$ in Eq.~\ref{eq:cae_layer_score} define probabilities over all candidate layers. We combine their mean probabilities with slot-normalized selection frequencies to obtain

\begin{equation}
 \begin{aligned}
 p_{i,r}^L
   &=\frac{\exp(b_i^{(r)})}
           {\sum_{k\in\mathcal{L}}\exp(b_i^{(k)})},\\
 f_r^L&=\frac{m_r}{nK_L},\\
 P_r^L&=\frac{1}{n}\sum_{i\in\mathcal{I}}p_{i,r}^L,\\
 \mathcal{L}_{\mathrm{bal}}^L
   &=N\sum_{r\in\mathcal{L}}f_r^L P_r^L.
 \end{aligned}
 \label{eq:supp_layer_balance}
\end{equation}
The selection frequencies satisfy $\sum_r f_r^L=1$.

\paragraph{Expert-level balancing.}
\begin{table}[t]
\centering
\caption{Real-world task procedures. $\dagger$ marks the long-horizon task.}
\label{tab:real_world_procedures}
\small
\renewcommand{\arraystretch}{1.05}
\begin{tabular}{@{}p{0.28\columnwidth}@{\hspace{0.7em}}p{\dimexpr0.72\columnwidth-0.7em\relax}@{}}
\toprule
\textbf{Task} & \textbf{Execution sequence} \\
\midrule
\raggedright Fruit collection & \raggedright Walk to table $\rightarrow$ pick up fruits $\rightarrow$ place in basket. \tabularnewline
\raggedright Bed making & \raggedright Bend down $\rightarrow$ grasp blanket corner $\rightarrow$ pull blanket into place $\rightarrow$ stand up. \tabularnewline
\raggedright Bag shelving & \raggedright Bend down $\rightarrow$ grasp bag $\rightarrow$ stand up $\rightarrow$ walk to shelf $\rightarrow$ place bag on shelf. \tabularnewline
\raggedright Table wiping & \raggedright Pick up cloth $\rightarrow$ wipe up spilled cola on the table while rotating torso and moving arm $\rightarrow$ put cloth back. \tabularnewline
\raggedright Cola dispensing & \raggedright Pick up cup $\rightarrow$ position under dispenser $\rightarrow$ fill with cola $\rightarrow$ pull cup back. \tabularnewline
\raggedright Laundry loading & \raggedright Open washer door $\rightarrow$ lift clothes from basket $\rightarrow$ step closer and load clothes $\rightarrow$ step back and close door. \tabularnewline
\raggedright Popcorn serving$^{\dagger}$ & \raggedright Grasp scoop $\rightarrow$ pick up bucket with other hand $\rightarrow$ scoop and pour popcorn into bucket $\rightarrow$ walk to table carrying bucket $\rightarrow$ set bucket securely on table. \tabularnewline
\bottomrule
\end{tabular}
\end{table}

Within each selected layer, balancing uses probabilities over all $E$ experts, conditioned on the tokens assigned to that layer. For $m_r>0$, we define

\begin{equation}
 \begin{aligned}
 p_{i,r,e}^E
   &=\frac{\exp(G_i^{(r,e)})}
           {\sum_{j=1}^{E}\exp(G_i^{(r,j)})},\\
 f_{r,e}^E&=\frac{c_{r,e}}{K_E m_r},\\
 P_{r,e}^E&=\frac{1}{m_r}
     \sum_{i\in\mathcal{I}}\delta_{ir}p_{i,r,e}^E.
 \end{aligned}
 \label{eq:supp_expert_statistics}
\end{equation}
The frequencies satisfy $\sum_e f_{r,e}^E=1$. Per-layer expert losses are averaged in proportion to the number of assigned tokens:

\begin{equation}
 \begin{aligned}
 \mathcal{L}_r^E&=E\sum_{e=1}^{E}f_{r,e}^E P_{r,e}^E,\\
 \mathcal{L}_{\mathrm{bal}}^E
   &=\frac{\sum_{r:m_r>0}m_r\mathcal{L}_r^E}
           {\sum_{r\in\mathcal{L}}m_r}.
 \end{aligned}
 \label{eq:supp_expert_balance}
\end{equation}
Layers with $m_r=0$ contribute zero. The losses encourage balance across the batch without requiring uniform routing for each token. We use $\lambda_L=\lambda_E=10^{-3}$ in Eq.~\ref{eq:action_grounding_loss}.

\paragraph{Selection frequency and fusion weight.}
Selection frequency measures how often a layer is included in the active set, whereas mean fusion weight measures its contribution to the projected features. For the valid token positions $\mathcal J$ of one observation and category, these quantities are

\begin{figure*}[!t]
\centering
\includegraphics[width=\textwidth]{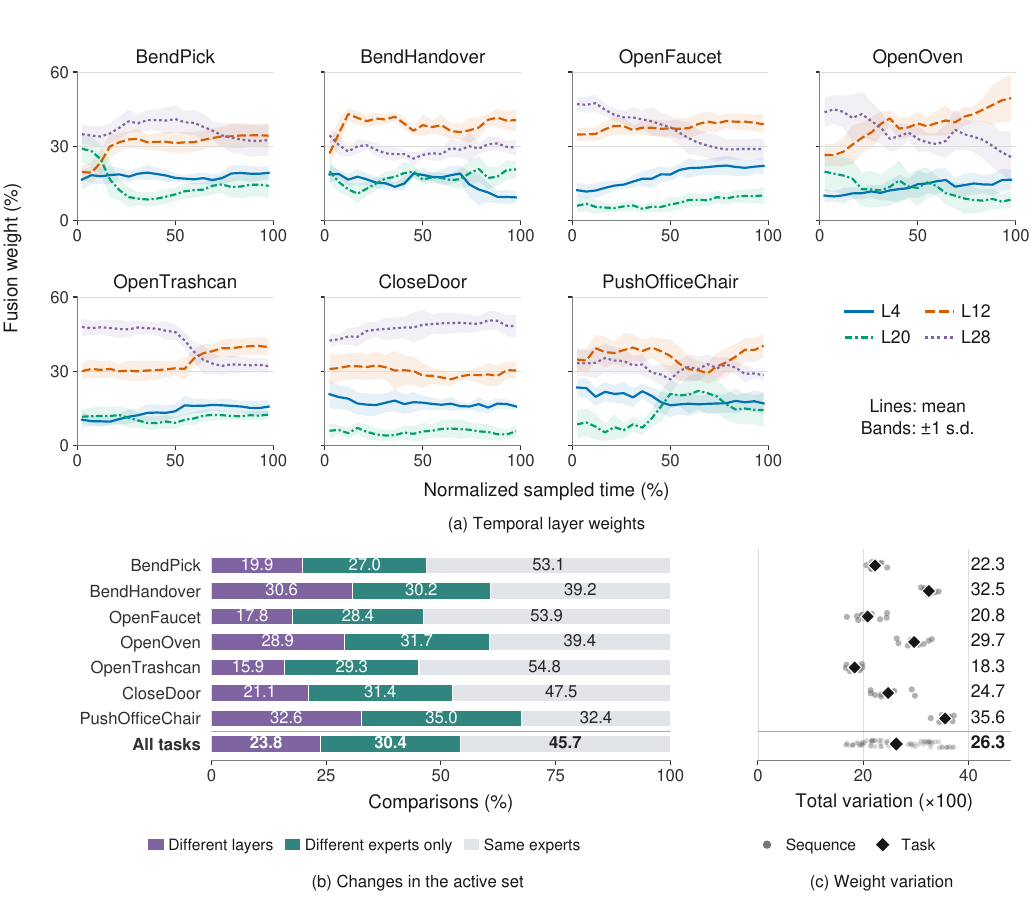}
\caption{\textbf{Temporal routing dynamics.} (a) Mean and standard deviation of image-token layer weights across sampled sequences for each task, with time normalized within each sequence. (b) Mutually exclusive changes in the active set between adjacent sampled observations at corresponding image-grid positions. ``Different experts only'' denotes a change of experts with the layer pair retained. (c) Total variation of the 32 expert weights ($\times100$), capturing changes in both selection and weighting. Circles denote sequence means, diamonds denote task means, and the final row pools all tasks.}
\label{fig:supp_cae_routing_temporal}
\label{fig:supp_cae_routing_changes}
\end{figure*}

\begingroup

\begin{equation}
\begin{aligned}
q_r &= \frac{1}{|\mathcal J|}\sum_{i\in\mathcal J}
 \delta_{ir},\\
\bar w_r &= \frac{1}{|\mathcal J|}\sum_{i\in\mathcal J}
 \sum_{e:(r,e)\in\mathcal S_i}\alpha_i^{(r,e)}.
\end{aligned}
\label{eq:supp_routing_measurements}
\end{equation}
The indicator $\delta_{ir}$ is defined above, and observation- and task-level averaging follows Sec.~\ref{supp:routing_analysis}. Because each token selects two layers, $\sum_r q_r=2$, while $\sum_r\bar w_r=1$. Expert-pair inclusion frequencies similarly sum to four, whereas their mean weights sum to one. The slot-normalized frequencies $f_r^L$ in Eq.~\ref{eq:supp_layer_balance} sum to one. Figures~\ref{fig:supp_cae_routing_profiles} and~\ref{fig:supp_cae_routing_temporal}a report fusion weights.

\endgroup

\section{Task Demonstrations and Evaluation Protocols}
\label{supp:protocols}
\label{supp:real_world_procedures}

Simulation and real-world demonstrations are collected through whole-body teleoperation of a Unitree G1 with Dex3 hands using a PICO headset and handheld controllers. Each episode records RGB observations, an instruction, measured joint states, and body and hand action targets. The state comprises 29 body-joint angles and 14 hand-joint angles; the action comprises a 64-dimensional SONIC motion latent and 14 hand targets.

\suppadd{For policy learning, episodes are represented on a 30 Hz timeline. RGB frames are paired with their state and action records by episode and frame index, with timestamps checked for alignment. A training sample uses the image and measured state at its observation frame, together with an action chunk starting at that frame. Image resizing and RGB caching preserve this temporal correspondence; augmentation is applied when the sample is loaded for training.}

Simulation evaluation comprises 10 trials at each of three randomization levels for each of seven tasks, yielding 30 trials per task and 210 per method. The execution sequences of the seven real-world tasks are summarized in Table~\ref{tab:real_world_procedures}. Each real-world task is evaluated over 10 trials, with success requiring completion without human intervention and without an imposed time limit. For both settings, we report per-task success rates and their mean across the seven tasks.

\section{Additional Results and Analyses}
\label{supp:results}
\subsection{Fixed-Layer Ablation Settings}
\paragraph{Fixed-layer architecture.}
The fixed-layer variants in Table~\ref{tab:simulation_ablations} extract a single decoder-block output at $r\in\{4,12,20,28\}$ and project it directly from $2048$ to $1536$ dimensions using one affine map. They retain CHAMP's six-block action DiT but omit the router and expert bank.

\paragraph{Initialization and training.}
All four variants use the VLM checkpoint obtained after $250{,}000$ motion-pretraining updates and keep it frozen during grounding and fine-tuning. Each variant undergoes $100{,}000$ grounding updates with a global batch size of $640$ and a constant learning rate of $10^{-4}$. Task fine-tuning initializes the linear projection and DiT from the corresponding layer-specific grounding checkpoint and resets the optimizer.

Each task is fine-tuned for $20{,}000$ updates with a global batch size of $128$ and a peak learning rate of $10^{-4}$, using $1000$ warmup updates followed by cosine decay. Other optimization settings follow Table~\ref{tab:supp_training_configuration}; action/state normalization and flow-time sampling follow Sec.~\ref{supp:flow_training}. The linear projection and DiT are trained solely with the flow-matching loss.

\subsection{CAE Routing Analysis}
\label{supp:routing_analysis}

\paragraph{Measurement protocol.}
We analyze seven separately fine-tuned task models using sampled observations.

Tokens are grouped by their positions in the multimodal sequence into image, instruction, and pose-text categories, excluding special tokens. Pose-text comprises wrist and pelvis information in the VLM prompt, distinct from the DiT's robot-state token. We average weights within each sampled observation and token category, then across samples and equally across the seven task models.

\paragraph{Layer and expert allocation.}
Routing profiles differ across token categories at both the layer and expert levels (Fig.~\ref{fig:supp_cae_routing_profiles}). Image tokens allocate 71.9\% of their weight to layers 12 and 28, while instruction tokens allocate 64.1\% to layer 4. Pose-text weights are more distributed, with layer 28 receiving the largest share at 39.8\%. Within layer 28, expert 5 receives 13.8\% of image-token weight and 2.8\% of pose-text weight, whereas expert 2 receives less than 0.1\% and 10.0\%, respectively.

\newpage
\paragraph{Temporal changes across samples.}
Routing also varies across temporally ordered samples within each task model (Fig.~\ref{fig:supp_cae_routing_temporal}a). Between the initial and final thirds of each sampled sequence, image-token weight on layer 28 decreases by 15.4 percentage points for OpenFaucet and increases by 4.6 for CloseDoor. In BendHandover, layer-28 weight remains nearly constant, while layer-4 weight decreases by 4.7 points and layer-20 weight increases by 4.1 points.

\paragraph{Route changes and their magnitude.}
We compare routing at the same 60 image-grid positions between temporally adjacent sampled observations. The selected layer pair changes in 23.8\% of comparisons; in 30.4\%, the layer pair is retained but the selected experts change; and in 45.7\%, the expert set is unchanged (Fig.~\ref{fig:supp_cae_routing_changes}b). We quantify weight variation by $0.5\|\mathbf w'-\mathbf w\|_1$, where each vector contains the weights of all 32 experts, with zeros for unselected experts. This measure lies in $[0,1]$, with task means ranging from 18.3\% to 35.6\% when expressed as percentages (Fig.~\ref{fig:supp_cae_routing_changes}c).

\end{document}